\documentclass{article}
\usepackage{ijcai22}

\usepackage{times}

\usepackage{soul}
\usepackage{url}
\usepackage[hidelinks]{hyperref}
\usepackage[utf8]{inputenc}
\usepackage[small]{caption}
\usepackage{graphicx}
\usepackage{amsmath}
\usepackage{booktabs}
\title{Neural Content Extraction for Poster Generation of Scientific Papers}

\author{
Sheng Xu \and
Xiaojun Wan
\\
\affiliations
Wangxuan Institute of Computer Technology, Peking University, Beijing, China \\
\emails
\{sheng.xu, wanxiaojun\}@pku.edu.cn
}

\begin{document}

\maketitle

\begin{abstract}
The problem of poster generation for scientific papers is under-investigated.
Posters often present the most important information of papers, and the task can be considered as a special form of document summarization.
Previous studies focus mainly on poster layout and panel composition, while neglecting the importance of content extraction.
Besides, their datasets are not publicly available, which hinders further research.
In this paper, we construct a benchmark dataset from scratch for this task.
Then we propose a three-step framework to tackle this task and focus on the content extraction step in this study. 
To get both textual and visual elements of a poster panel, a neural extractive model is proposed to extract text, figures and tables of a paper section simultaneously.
We conduct experiments on the dataset and also perform ablation study.
Results demonstrate the efficacy of our proposed model.
The dataset and code will be released.
\end{abstract}


\section{Introduction}

Many researchers rely on posters to present their research work in an effective and expressive way.
However, it takes much effort to make posters from papers.
Given a certain paper, one needs to extract the most important points and then consider the layout of the poster so that the information is presented clearly.
It would be more than helpful if drafting posters can be generated from papers automatically.
Researchers only need to make minor modification to get the final poster, which will save much time.

The problem of automatic poster generation is far under-investigated by now.
Posters summarize the most important points of papers and also arrange their positions properly.
The space of a poster can often be divided into several panels, and each panel focuses on its own aspect respectively.
Apart from text, posters also include graphical elements like figures and tables.
In the following, we will use graphs to collectively refer to figures and tables in the paper.
More often than not, people extract sentences, tables and figures from papers, and then make slight modification to make posters.
To generate posters automatically, the system needs to extract not only text, but also graphs related to text as well.
What's more, the system needs to decide the layout of poster and composition between the panels.
The layout should be tidy and well-designed so that the poster can attract more attention.

Previous work mainly focuses on layout and composition of posters~\cite{paramita2016tailored,qiang2016learning,qiang2019learning}, while the problem of content extraction is always ignored.
They employ naive models to handle this part, which do not perform well. 
We argue that content extraction should be paid more attention to.
After all, 
what really matters and attracts more attention from readers is the content of the poster.
So in this study, we concentrate more on the problem of content extraction.
Previous work hypothesizes that each section from the paper is reflected in the poster as a panel.
Although this hypothesis applies to some cases, it is common to see sections which are not aligned to any panels.
This may be because they are not important enough or their main points are covered by other sections.
Moreover, many posters generated by previous work ignore graphs and are made of text only.
These posters may bore readers with a bunch of text and no graphical elements.
Although a few studies were done on poster generation, to our knowledge,
a benchmark dataset has not been developed so far,
which hinders further research on this topic.

In this study, we first collect pairs of papers and posters from public conference web pages,
and then build a dataset which explicitly aligns sections of a paper to panels of its corresponding poster.
Then we try to tackle the problem of automatic poster generation in three steps.
As mentioned above, some sections are not important enough, so we filter them out at step one.
Then at step two, given a certain section, we propose a neural model to extract important sentences and graphs to form a poster panel.
We capture the reference relationship between sentences and graphs in a section to assist the extraction process.
At step three, all panels generated during step two are used to form a final poster.
We pay more attention on step two because the problem of content extraction is very important for poster generation while it has received less attention in previous work. 
We implement several baselines for comparison and conduct ablation study to further understand the influence of model components and data.
The experimental results show the effectiveness of our model.

The contributions of this paper can be summarized as follows:
\begin{itemize}
    \item We create a dataset for the task of poster generation of scientific papers. The dataset is ready to be released for research purpose.
    \item We present a three-step method to generate posters. Especially, we focus on content extraction from papers and propose a neural extractive model to extract important sentences and graphs at the same time. 
    \item Automatic and human evaluation results show the efficacy of our proposed method. We also adopt ablation study to verify the usefulness of model components. 
\end{itemize}


\section{Related Work}

Posters show the most important points of scientific papers.
Hence poster generation can be seen as a special form of document summarization.
Recently, neural network models have been more and more popular with both abstractive summarization ~\cite{chopra2016abstractive,im2021self} and extractive summarization \cite{nallapati2017summarunner,jiadeep} on account of their excellent performance.
However, the research on automatic poster generation is limited.
To handle this problem, Paramita et al~\cite{paramita2016tailored} make use of models like Naive Bayes to classify sentences of a paper into several categories based on hand-crafted features.
They group the sentences into several classes and then fill them into templates to get the final result.
However, their model makes use of hand-crafted features only and is no match for neural network based methods.
What's more, they totally ignore the graphical elements of the paper, and the poster is composed of text only.

The work of ~\cite{qiang2016learning,qiang2019learning} focuses mainly on poster composition.
They use hand-crafted features of the section to predict attributes for the panel, like its size and aspect ratio.
After that, a recursive page splitting algorithm is proposed to generate panel layout.
However, they extract the sentences with TextRank model, which is unsupervised and the performance is not quite satisfying.
The system also requires human to manually select figures and tables which are used in the poster.
What's more, their method is based on a strong hypothesis that each section in the paper corresponds to one panel in the poster, and that each subsection corresponds to one bullet point in the panel, which is not realistic.
Finally, the dataset they conduct experiments on consists of only 85 paper-poster pairs and cannot be accessed by public.

Slides are another common way to present scientific papers.
There are several works conducted on the topic of generating slides from papers~\cite{hu2013ppsgen,wang2017phrase,sefid2019automatic,fu2021doc2ppt,li2021towards}.
Slides are also considered to be special summaries of papers, and many researchers make use of summarization models to tackle the problem.
The difference between slides and posters is that slides consist of multiple pages, and pay more attention on hierarchical structure of the points.
They can also use animations.
While posters are static and organize the content in the form of panels.

Poster generation is related to multimodal summarization ~\cite{zhu2018msmo,chen2018abstractive}.
They both need to sum up the most important parts from the document and present them with both text and images. However, multimodal summarization is usually focused on news or wikipedia domains. Moreover, the natural images in these domains are relatively easier to understand than figures/tables in scholarly papers and thus existing multimodal summarization methods usually use image features directly. 


\section{Dataset}

For now, there is no publicly available dataset for poster generation of scientific papers.
Thus we aim to construct a benchmark dataset for this task.
We crawl the homepages of several academic conferences.
Some of the accepted papers provide posters and other additional materials which allow further modification and usage for research purposes. We download and collect these paper-poster pairs as raw data.
Most posters are in PDF format, while some are represented in PPT format.
The text of each paper is extracted manually \footnote{This will not hinder the use of our proposed system for poster generation in practice, because the paper texts can be easily obtained from the authoring tools used by authors such as LaTeX and Word.} because we have tried several existing automatic PDF processing tools but found they did not work well.
As for extraction of figures, tables and their captions, we use a tool named pdffigure $\footnote{https://github.com/allenai/pdffigures}$, which performs generally well.
After automatic extraction of graphs and captions is done, we examine and correct the results manually should there be any mistakes.

The next step is alignment.
As mentioned above, each panel in the poster focuses on its own aspect, and can be aligned to a certain section in the paper more often than not.
We employ human aligners to explicitly label the alignment relation.
Human aligners are asked to read the paper and then align the panels of a poster to sections of its corresponding paper according to their content.
Many people directly use tables and figures from the paper when making its poster, although some may make minor adjustment.
For automatic poster generation, we follow this extractive manner and further ask the aligners to judge whether each table or figure in the poster originates from the paper.
The tables and figures of the paper which are aligned to the poster are labeled positive.
Each paper-poster data sample is represented to and aligned by two aligners.
If their alignment disagrees with each other, a third aligner will check the results once more and make the final decision.
The aligners are university students who are familiar with research and good at English.

Finally, we get a dataset consisting of 260 poster-paper pairs and 846 section-panel pairs.
The agreement rate of section-panel alignment is 82.15\%.
For training content extraction models, we follow the work of \cite{zhou2018neural} to label each sentence as positive (i.e., it should be extracted into the panel text) or not by greedily optimizing ROUGE-2 F1 on gold-standard panel text.
The labels of graphs are obtained from human aligners.
Finally, each section-panel data sample is composed of the following elements:
\begin{itemize}
    \item Sentences of the section and their labels indicating if they should be extracted or not.
    \item Figures and tables included in the section, together with their captions and labels. Some data samples may not contain these, because those sections may not contain any figure or table.
    \item Text, figures and tables of the panel.
\end{itemize}

We randomly partition the samples into 10 groups and apply 10-fold cross validation in our experiments.
At each time, the model takes eight groups for training, one group for validation and the rest one for testing.
The data samples originate from same paper are partitioned into the same group to avoid overlap between different groups.
In Table \ref{tab:dataset-stat} we report some descriptive statistics of our dataset.
About 55.34\% of the paper graphs are used to in the poster.
The overall text compression ratio is 16.87.
The large compression ratio also shows that posters highly summarize the content in papers.

\begin{table}[ht]
    \centering
    \begin{tabular}{lr}
        \hline
        \# section-panel pair & 846 \\
        \# graph & 2221 \\
        \# positive graph & 1229 \\
        section text length & 692.77 \\
        panel text length & 41.07 \\
        \hline
    \end{tabular}
    \caption{Statistics of our dataset. Almost half of the tables and figures are not chosen to be presented in the posters. The average text length of poster panels is much less than that of the aligned sections. }
    \label{tab:dataset-stat}
\end{table}


\section{Methodology}

\begin{figure}
    \centering
    \includegraphics[width=0.4\textwidth]{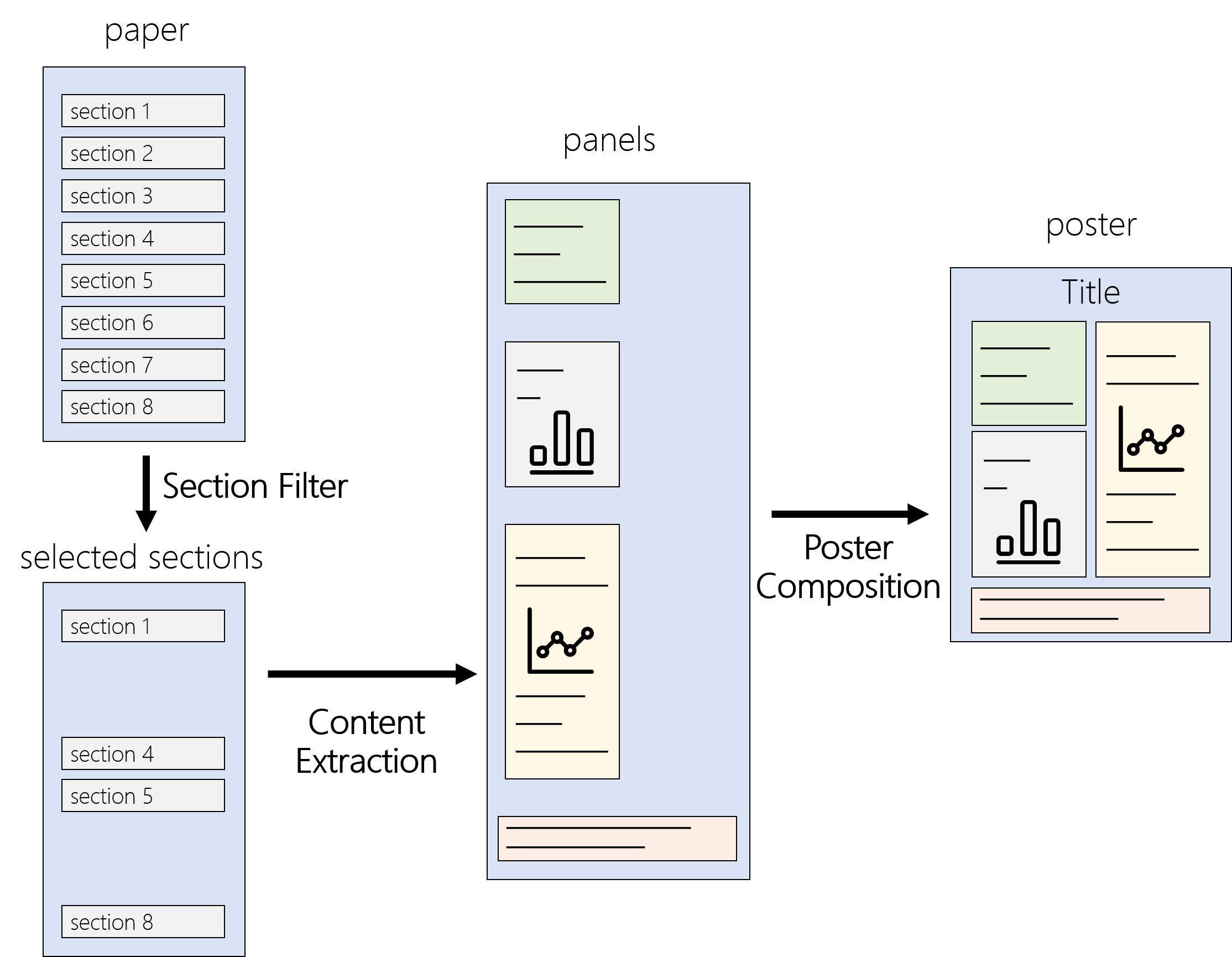}
    \caption{The three-step method for poster generation. Given a paper, we first identify important sections. Then the important sections are used to generate panels. Finally we use the panels to compose the poster.}
    \label{fig:method-framework}
\end{figure}

Given a certain paper, we propose to generate its poster with a three-step method, as shown in Figure \ref{fig:method-framework}.
First, we use a section filter model to learn each section's importance with respect to the whole paper, and then filter out the unimportant sections.
Then the most important sentences and graphs of the remaining sections are extracted by using a content extraction model, which is the focus of this work.
At last, the poster composition model produces the poster with the sentences and graphs extracted from the sections.
In the following sections, we will briefly describe the section filter step and the poster composition step, while describe the content extraction step in detail.

\subsection{Section Filter} \label{sectionfilter}

As stated above, not all sections from a paper should be selected to make panels when making a poster.
Some sections are not important enough, and it is unnecessary to include these sections in the poster due to the limited space.
Thus we need to first filter some certain sections out.

Specifically, given a paper, we first encode the text of each section separately. We utilize a pre-trained model RoBERTa\cite{liu2019roberta}  for encoding, and it returns a vector representation of each section.
We consider the vectors of sections as a sequence, and further use a transformer encoder to obtain their paper-level representations.
After that, a classifier is used to evaluate each section's importance degree and decide whether to choose it as one of the source sections of poster.

In the experiments, we compare our model with the following two ablation models:
\begin{itemize}
    \item \textbf{Ablation Model 1}: We want to verify whether the content text of sections is necessary to evaluate their importance. The model is the same as the one described above while it takes only the titles of sections as input.
    \item \textbf{Ablation Model 2}: This model does not use the transformer encoder. After using RoBERTa to get the vector representation of a section, the model directly uses a classifier to predict its label without learning and using its paper-level representation.
\end{itemize}

The section classification accuracy of our model, ablation model 1 and ablation model 2 is 88.01\%, 74.52\%, and 84.08\% respectively.
From the result, we notice that the titles of sections are not sufficient to determine their importance.
It is the content texts of sections that matter.
We also learn that the global information of the whole paper is beneficial for learning representations of sections.



\subsection{Content Extraction}

Given a section from the paper which includes a few sentences and may additionally contain tables and figures, 
we want to select the most important sentences, together with the tables and figures which are highly relevant to selected sentences. Here we decide not to directly use the images of figures and tables but to utilize their captions instead. Usually the pictorial elements in figures are abstract and hard for models to understand, and models can be poor at learning the information revealed in the tables with a bunch of numbers.
Fortunately, most tables and figures in papers are accompanied with captions which describe and summarize them in natural language texts.
Thus it is easier for models to learn the meanings of tables and figures from their captions.

Formally we define the content extraction problem as follows:
Given sentences from a section: $\{s_1, s_2, ... s_n\}$, together with the captions of the graphs (i.e., figures and tables) included in the section: $\{c_1, c_2, ... c_m\}$, the model needs to extract the most important sentences and graphs.

The tackle the problem, we propose a model which learns to predict the importance scores of both sentences and graphs simultaneously. The idea underlying our model is that the extraction of sentences and the extraction of graphs can help each other. 
We further make use of additional attention weight to help the model capture the relationship between sentences and graphs.
This turns out to help stabilize and improve the performance of our model.

Specifically, we first use a pretrained RoBERTa model to encode both sentences and graph captions in a section.
Because every caption starts with word ``Figure" or ``Table" which is followed by its number, the model can easily distinguish captions from sentences after training.
Thus we do not employ additional mechanism to differentiate these two kinds of source inputs.
All the input sentences and captions are encoded separately at this stage.
We get vector representations $W=\{w_1, w_2, ... w_n\}$ and $U=\{u_1, u_2, ... u_m\}$ for section sentences and graph captions respectively: $w_i=RoBERTa(s_i), u_j=RoBERTa(c_j)$.

We assume the context information of a section matters a lot to evaluate the importance of each sentence or graph in the section.  Moreover, the captions can also be used to help learn the importance of sentences.
If a sentence is semantically related to an important graph, it is highly possible that the sentence is also significant, and vice versa.
Therefore, we add transformer encoder \cite{vaswani2017attention} layers stacked on top of RoBERTa outputs, i.e., the RoBERTa outputs for section sentences and graph captions are concatenated together as input for the transformer encoder. 
These layers encode both sentences and graph captions to learn their section-aware representations.

We innovatively add additional attention weight during the calculation of self-attention, in order to help the encoder catch the relationship between sentences and graphs better.
Recall that the original transformer encoder calculates attention matrix following the equation below:

\begin{equation} \label{eq:attention}
    Attention(Q, K)=softmax(\frac{Q K^T}{\sqrt{d_k}})
\end{equation}

\noindent $d_k$ is the dimension of the vectors.
$Q$ and $K$ are query and key matrices respectively, and they are calculated from document matrix, which in this case is the concatenation of sentence matrix and caption matrix $D=\{W, U\} ^ T = \{w_1, w_2, ...w_n, u_1, u_2, ... u_m\} ^ T$, $Q=D \cdot W_Q , K = D \cdot W_K$.
The additional attention weight matrix $A$ of our model is a square matrix of order $n+m$, and it is built in the following way.
Originally, all elements in the matrix are all set to be 0.
For a certain graph $c_i$, we find all sentences that refer to it.
Suppose the indices of these sentences are $J=\{j_1, j_2, ... j_t\}$.
These sentences are considered more relevant to the graph.
Hence we increase the attention weight between these sentences and the graph caption.
Specifically, we add $h_1 / t$ to $A_{n+i, j}$ and $A_{j, n+i}$, for all $j \in J$.
The sentences that refer to the same graph are also regarded relevant, and we further add $h_2 / t$ to $A_{j,k}$, for all $j,k \in J, j \neq k$.
$h_1$ and $h_2$ here are hyper-parameters and $t$ is the size of $J$. $t$ is used as a normalization factor for $h_1$ and $h_2$. 
After all graphs are visited, we finally get the additional attention weight matrix.
Our model calculates attention matrix in the following new way:

\begin{equation} \label{eq:new-attention}
    Attention(Q, K)=softmax(\frac{Q K^T}{\sqrt{d_k}} + A)
\end{equation}

The remaining parts of our encoder, such as feed-forward network, LayerNorm and residual connection, are the same as the original transformer encoder.
After that, we use fully connected layers to predict the importance scores of the sentences and graphs $P=\{p_1, p_2, ... p_{n+m}\}$.

Note that the number of negative sentences are overwhelmingly larger than that of positive ones. In addition, there are more training samples for sentence extraction than that for graph extraction.
In order to deal with the issue of imbalanced data distribution, we make use of balanced cross entropy loss:
\begin{equation}
    loss_{sentence/graph} = \left\{
    \begin{aligned}
        -\alpha log(p), \text{ if } y = 1 \\
        -\beta log(1-p), \text{ if } y = 0
    \end{aligned}
    \right.
\end{equation}
$y$ is the gold label of one sentence or graph, $p$ is the importance score predicted by the model, and
$\alpha$ and $\beta$ are hyper-parameters that control the importance of different parts.
Finally, our model optimizes the weighted sum of sentence loss and graph loss simultaneously: $ loss=loss_{sentence}+\gamma * loss_{graph} $.
$\gamma$ is set to be larger because the numbers of graph captions are less than section sentences.

\subsection{Poster Composition}

After getting contents of all panels, including text and corresponding graphs, the final step is to integrate the panels to get the poster.
Previous work generates posters in portrait format only.
However, 
some conferences may demand the posters to be certain format.
We decide that the users should choose whether the poster is landscape or portrait format.

Following the work of \cite{paramita2016tailored}, we predefine several templates of both portrait and landscape formats.
The model will search for proper templates based on the format user chooses, the numbers of panels, text length, graph number, graph size and other attributes.
The output of the model is a LaTeX document which makes use of tikzposter package.
It generates scientific poster after compilation.
People can make further modification
to get customized poster.


\section{Evaluation Setup}

Note that the content extraction step is the focus of this study and in this section we implement several baselines models for comparison in either text extraction or graph extraction. 
We also perform ablation study to further understand the influence of different factors on our model.

\subsection{Implementation Details}
Specifically, the encoder is composed of 3 identical layers.
The dimensionality of input and output of encoder layers is $d_k = 768$, the dimensionality of feed-forward inner-layer is $d_{ff}=1536$.
$h_1$ and $h_2$ are set $1e-2$ and $1e-3$ separately for all layers.
During training, the weights of $loss_{sentence}$ are set $\alpha = 1.0, \beta = 0.5$, while for $loss_{graph}$, $\alpha = \beta = 1.0$, and $\gamma$ is set 3.0.
We use Adam optimizer with an initial learning rate $lr=8e-5$, momentum $\beta_1=0.9$, $\beta_2=0.999$.

\subsection{Compared Models for Text Extraction}

The following four extractive models and one abstractive model are used for comparison: 

\textbf{Lead-3}: Lead-3 is an extractive baseline which directly chooses and concatenates the first three sentences of source document to get a summary.

\textbf{TextRank}: TextRank is an unsupervised method to calculate the importance of the sentences based on their similarity ~\cite{mihalcea2004textrank}.
After applying TextRank algorithm, the model greedily selects the most important sentences to get summary until the total length reaches the limit.
This is also the method used in \cite{qiang2016learning,qiang2019learning} for sentence extraction.

\textbf{BERTSum}: BERTSum is a simple variant of BERT for extractive summarization \cite{liu2019fine}.
After concatenating all sentences of input document together, it uses pre-trained BERT model ~\cite{devlin2018bert} for encoding.
The model further builds several layers on top of BERT outputs to capture document-level features, and then calculates importance scores of the sentences.
Apart from BERTSum, we also try \textbf{RoBERTaSum}, which replaces BERT model with RoBERTa, because RoBERTa generally performs better than BERT.

\textbf{GPT-2}: Other than the extractive summarization models, we finetune GPT-2 ~\cite{radford2019language} on this task. 
The source input is the text of a section, and its target output is the text of the corresponding panel. This model is a strong abstractive model. 

We also implement three ablation models for comparison: 

\textbf{Without weight}\label{woweight}: In this ablation model, we discard the additional attention weight and follow Equation \ref{eq:attention}, i.e., using the original transformer encoder to calculate attention weight.

\textbf{Without caption}: In this model, the graph captions in the data are all abandoned. The model only uses the sentences of the section and predicts their importance scores after learning their section-aware representations.


\textbf{Image}: We borrow the idea from multimodal summarization and come up with this baseline.
The baseline is the same as our model except that it extracts graph features by utilizing ResNet~\cite{he2016deep} while our model manages that by using a RoBERTa to encode graph captions.

\textbf{Oracle}: Finally we implement the extractive oracle model. It extracts sentences from source document to optimize ROUGE-2 F1 score.
The performance of oracle is considered as the upper bound of extractive models.

\subsection{Compared Models for Graph Extraction}
Two baseline models and three ablation models are used for comparing graph extraction results:

\textbf{Similarity}:
This model is simply based on textual similarity and the score of a graph is the maximum cosine similarity between the caption and each sentence in the section.
We decide a threshold value based on training data.
The graphs with scores larger than the threshold are labeled positive.

\textbf{RoBERTa}: For comparison, we finetune a RoBERTa model to classify graphs.
The model encodes the caption and then use a classifier to decide whether to choose that table or figure without using the section's text.

\textbf{Without section}: The ablation model abandons all section sentences and only uses captions. 


In addition, we also use \textbf{Without weight} and \textbf{Image} models, as described in the prior subsection.

\section{Evaluation Results}

\subsection{Automatic Evaluation}

Note that we use 10-fold cross-validation in the content extraction experiments, we report the mean and standard derivation of the 10 experimental results.
For text extraction part, we follow the previous work of document summarization and report ROUGE scores~\cite{lin2004rouge} in Table \ref{tab:experimental-result-rouge}.

\begin{table}[ht]
    \centering
    \resizebox{\linewidth}{!}{
    
    \begin{tabular}{l|ccc}
        \hline
        model &  ROUGE-1 & ROUGE-2 & ROUGE-L \\
        \hline
        
        Lead-3 & $19.29 \pm 2.00$ & $5.95 \pm 1.60$ & $13.49 \pm 1.62$  \\
        TextRank & $20.88 \pm 1.36$ & $6.69 \pm 5.90$ & $13.87 \pm 1.00$  \\
        BERTSum & $23.56 \pm 1.25$ & $6.47 \pm 0.86$ & $15.32 \pm 0.83$ \\
        RoBERTaSum & $23.79 \pm 1.18$ & $6.76 \pm 0.92$ & $15.60 \pm 0.76$  \\
        GPT-2 & $18.42 \pm 1.72$ & $7.10 \pm 1.30$ & $14.70 \pm 1.49$  \\

        \hline
        
        our's & $\textbf{24.74} \pm 0.60$ & $\textbf{7.81} \pm 0.81$ & $\textbf{16.50} \pm 0.63$ \\
        w/o weight & $24.69 \pm 0.87$ & $7.50 \pm 1.04$ & $16.25 \pm 1.01$ \\
        w/o cap & $24.16 \pm 1.46$ & $7.03 \pm 0.98$ & $15.98 \pm 0.96$  \\
        Image & $24.11 \pm 1.20$ & $7.19 \pm 1.27$ & $16.07 \pm 1.00$ \\

        \hline
        Oracle & $41.36 \pm 2.33$ & $22.94 \pm 2.27$ & $25.59 \pm 1.96$  \\
        \hline

    \end{tabular}
    
    }
    \caption{ROUGE scores for text extraction in the Content Extraction step. Note that GPT-2 is an abstractive model.}
    \label{tab:experimental-result-rouge}
\end{table}

As illustrated in the table, our proposed model outperforms all baseline models.
From the performance of Lead-3, we notice that the first several sentences can still cover some of the important points of a section.
TextRank performs a little better than Lead-3.
RoBERTaSum gets better scores than BERTSum on all three ROUGE metrics, 
and they generally perform better than Lead-3 and TextRank.
It can be observed that the performance of GPT-2 is relatively low, as compared with the two neural extractive summarization models.  
The size of the training data may be still not large enough to finetune a generation model with huge parameters.

The difference between RoBERTaSum and the ablation model \textbf{without caption} lies in the way they encode the section text: the former encodes the whole text after concatenating all sentences, while the latter encodes all sentences separately.
As shown in Table \ref{tab:dataset-stat}, the text length of section can easily reach the length limitation of BERT, which is 512.
By encoding the section as a whole, 
truncation is required and can lead to information loss.
As a result, RoBERTaSum performs a little lower than \textbf{without caption}.
While our full model further outperforms \textbf{without caption}, indicating that the graph part can also benefit text extraction.
The performance of \textbf{Image} is close to \textbf{without caption}, which shows that the graph features extracted from images cannot help much.


The ablation model \textbf{without weight} performs slightly lower than our full model.
It seems that the model can learn a little bit better when the attention weight matrix between sentences and graphs is given explicitly.
We also observe that the standard derivation of our model is lower than that of the ablation models, which indicates that with the help of the attention weight matrix, our model can perform more stably.

\begin{table}[ht]
    \centering
    \small
    \begin{tabular}{l|c}
        \hline
        model & graph acc. (\%) \\
        \hline
        Similarity &  $63.02 \pm 5.39$ \\
        RoBERTa &  $65.75 \pm 5.05$ \\
        \hline
        our's & $\textbf{68.17} \pm 5.79$ \\
        w/o weight & $67.35 \pm 6.33$ \\
        w/o sec &  $66.07 \pm 5.21$ \\
        Image & $60.02 \pm 4.65$ \\
        \hline
    \end{tabular}
    \caption{Graph extraction accuracy in the Content Extraction step.}
    \label{tab:experimental-result-acc}
\end{table}

The accuracy of the graph extraction models is reported in Table \ref{tab:experimental-result-acc}.
We can see that the accuracy of similarity based model reaches a relatively high level, which means the textual similarity does help decide whether the table or figure is important.
The accuracy of RoBERTa is a little higher than similarity based model, although it makes no use of section text.
We can draw a conclusion that section sentences can help the model better capture the features of graphs when comparing the performance of our full model with ablation model \textbf{without section}.
The performance of \textbf{Image} is relatively lower, which indicates that the model performs poorly at learning from their images.

% ...
From the experimental results, we notice that
the part of text extraction and graph extraction can benefit from each other during training because they are closely connected.
By learning and predicting the importance of sentences and graphs at the same time, our model performs better on both parts than baseline models.

\subsection{Human Evaluation}
Human evaluation is also conducted for text extraction.
We ask human judges to assess the output text of our model against other baseline models, namely TextRank, RoBERTaSum and GPT-2.
The evaluation is conducted in three aspects:
\begin{enumerate}
    \item Informativeness: whether the output text covers important points of the section.
    \item Fluency: whether the output text is well-formed and grammatically correct.
    \item Non-Redundancy: whether the output text contains redundant or repeated information.
\end{enumerate}
For each input sample, people are asked to judge whether the output of our model outperforms the specific baseline in the three aspects. 
Their answers are chosen from win, lose and tie and they are not told which one of the pair is the output of our model.
We randomly choose 40 samples and get 120 pairs totally.
Each output pair is represented to two judges.
Judges are university students who are good at English and familiar with the research field of that paper. The exact match ratio of human annotation is 66.94\%.

\begin{figure}
    \centering
    \includegraphics[width=0.43\textwidth]{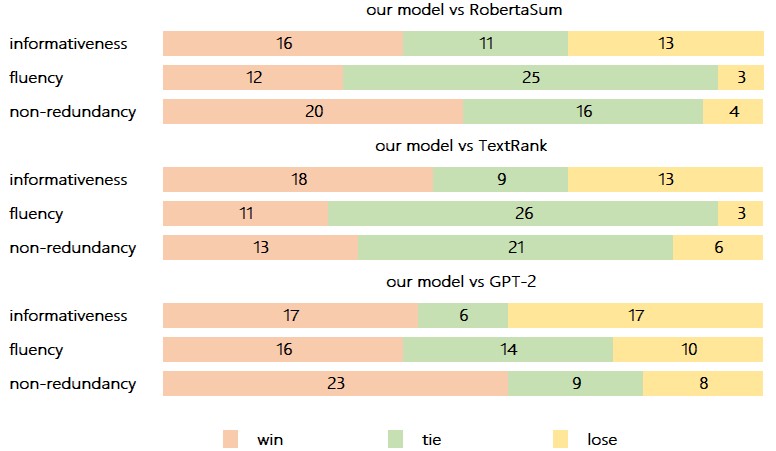}
    \caption{Human evaluation results for text extraction.}
    \label{fig:human-eval-step2}
\end{figure}

\begin{table}[]
    \centering
    \tiny
    \begin{tabular}{|p{0.45\textwidth}|}
    \hline
        \textbf{Human}: transformer performance is highly dependent on batch size (multi-gpu systems and gradient accumulation are very helpful). domain adaptation through fine-tuning provides improvements. adding noise helps to take advantage of more back-translations.
        further study is needed on effects of noise jointly with fine-tuning. \\
    \hline
        \textbf{Our's}: the results show that the performance of models following the transformer architecture is highly dependent on the batch size used to train the model, requiring multiple gpus or gradient accumulation in order to fully take advantage of this architecture. \\
    \hline
        \textbf{RoBERTaSum}: the experiments carried out this year have allowed us to explore one of the missing pieces of our wmt18 submission, which is the interaction between the transformer architecture and different batch sizes. \\
    \hline
    \end{tabular}
    \caption{Examples of generated panel texts.}
    \label{tab:case-study}
\end{table}

The results are presented in Figure \ref{fig:human-eval-step2}.
Our model performs better than or as well as other baseline models on all three aspects.
The results indicate that our model can produce panel texts which are more fluent and less redundant.




\subsection{Case Study}

In Table \ref{tab:case-study}, we present examples of panel texts produced by our model and RoBERTaSum.
The ground truth focuses on different factors that affect the performance of transformer model.
The output of our model also states the influence of  batch size, multiple GPUs and gradient accumulation, while still missing some points due to length limitation.
However, RoBERTaSum only extracts a transition sentence, which is not quite important.

\begin{figure}
    \centering
    \includegraphics[width=0.33\textwidth]{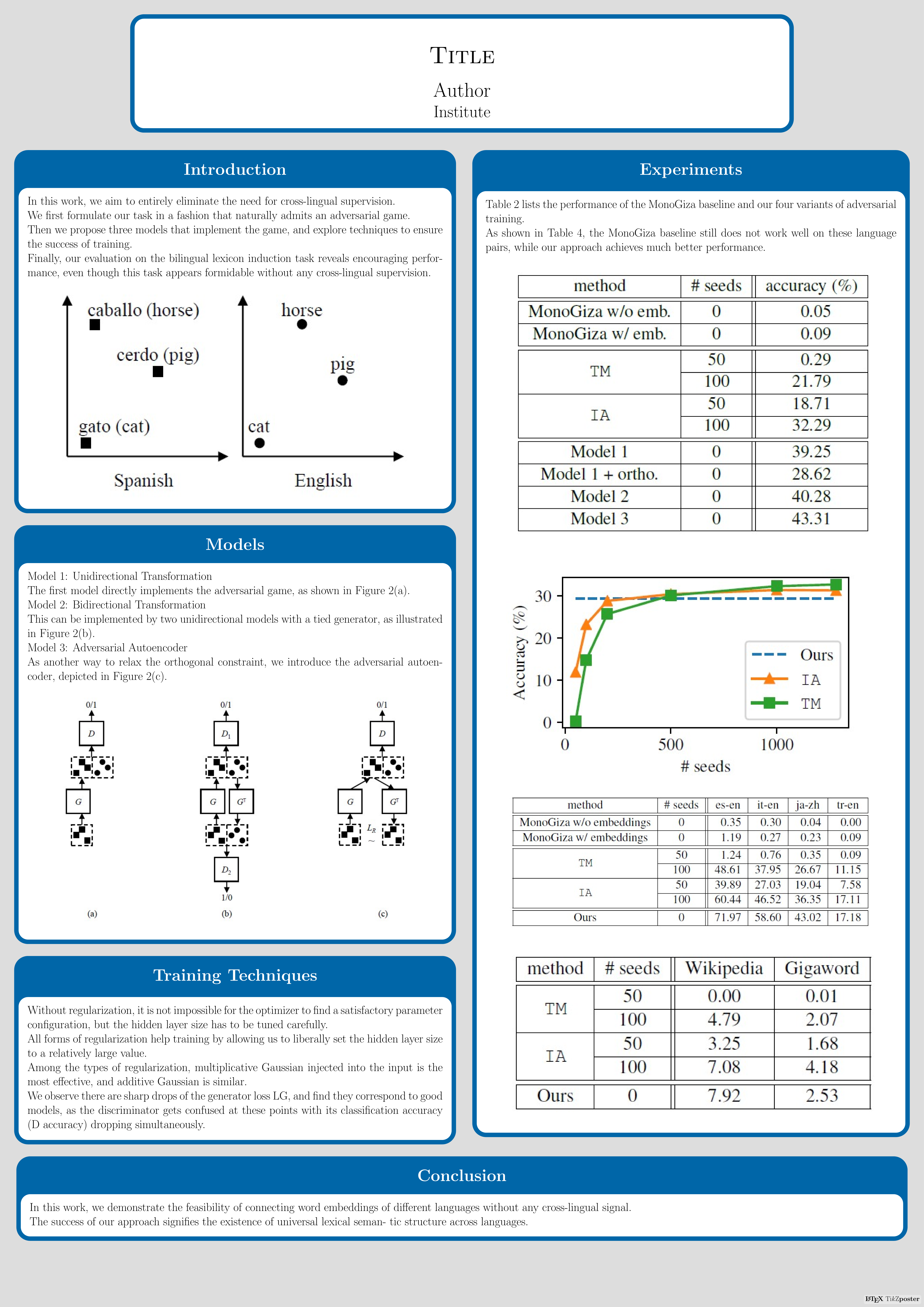}
    \caption{An poster generated by our method.}
    \label{fig:case-study-poster}
\end{figure}

In Figure \ref{fig:case-study-poster}, we present a poster generated by our three-step method.
As shown in the poster, our model manages to extract important points together with related graphs.
The panels are well-placed and the layout is reasonable.






\section{Conclusion}

In this study, we build a dataset on automatic poster generation and design a three-step method in order to overcome the shortcomings of previous work. Particularly, we propose a neural content extraction model to extract both sentences and graphs at the same time. 
The evaluation results verify the effectiveness of our model.
For future work, we will further enlarge out dataset. We will also try to merge the three steps and tackle this problem with one end-to-end model.


\bibliographystyle{named}
\bibliography{ijcai22}

\end{document}